\documentclass[runningheads]{llncs}

\usepackage{hyperref}
\usepackage[T1]{fontenc}
\usepackage{amsmath}

\usepackage{multirow}
\usepackage{booktabs} 
\usepackage{subcaption}
\usepackage{soul}
\usepackage{xcolor}
\soulregister{\cite}7

\usepackage{graphicx}
\begin{document}
\title{
Improving Multi-label Recognition using Class Co-Occurrence Probabilities}


\titlerunning{Improving MLR using Class Co-Occurrence Probabilities}
%
\author{
    Samyak Rawlekar \inst{1} \thanks{Equal contribution.} \and
    Shubhang Bhatnagar \inst{1} \textsuperscript{*} \and
    Vishnuvardhan Pogunulu Srinivasulu \inst{2} \and
    Narendra Ahuja \inst{1}
}

\authorrunning{Rawlekar et al.}
%
\institute{University of Illinois Urbana-Champaign, IL, USA
\email{\{samyakr2,sb56,n-ahuja\}@illinois.edu}
\and
Vizzhy, USA
\email{vishnu@vizzhy.com}
}

\maketitle              
\begin{abstract}
Multi-label Recognition (MLR) involves the identification of multiple objects within an image. To address the additional complexity of this problem, recent works have leveraged information from vision-language models (VLMs) trained on large text-image datasets for the task. These methods learn an independent classifier for each object (class), overlooking correlations in their occurrences. Such co-occurrences can be captured from the training data as conditional probabilities between a pair of classes. We propose a framework to extend the independent classifiers by incorporating the co-occurrence information for object pairs to improve the performance of independent classifiers. We use a Graph Convolutional Network (GCN) to enforce the conditional probabilities between classes, by refining the initial estimates derived from image and text sources obtained using VLMs. We validate our method on four MLR datasets, where our approach outperforms all state-of-the-art methods.

\end{abstract}
\section{Introduction}
\label{sec:intro}

Multi-label recognition (MLR) involves identifying each of the multiple classes from which objects are present in an image. It has many applications such as identifying all different: diseases evident in a chest x-ray  \cite{huang2024radiology}, products in a query image for e-commerce \cite{ecomm_product}, and food items in a plate for diet monitoring systems \cite{nutrient3,nutrient2}.
MLR is more challenging than classifying images having a single object \cite{deng2009imagenet,dosovitskiy2020image} because an image may contain combinatorially large mix of classes, for which learning would require exponentially larger number of images (O($2^{N}$) images for N classes) than for Single label recognition (SLR). The objects may also occur in different layouts so recognition either requires object segmentation and recognition of each segmented object independently, or recognizing the object mix from the features measured over the entire image.

Several approaches \cite{dualcoop} have taken the latter approach, namely recognition from image level features. Further, these approaches also reduce complexity by recognizing the presence of each object in an image independent of the others that may also be present. Training here amounts to learning independent classifiers for each object, which are then used to detect the corresponding objects, both using image level features. Apart from not segmenting the image and features into those for different objects, these methods also neglect the evidence for the presence of an object provided by the context of other objects. In practice, many objects are in sets, making their occurrences interdependent. Using independent classifiers neglects the mutual information present, which, if used, could enhance the performance of individual classifiers. This is particularly important in view of the already larger amount of data needed for MLR, and the relatively small sizes (vs SLR) of available annotated MLR datasets (because multiple classes present in an image require more annotation effort than required for SL images).

 To mitigate the paucity of labeled data, recent MLR approaches \cite{scpnet,dualcoop} have focused on adapting large Vision Language Models (VLMs) e.g. CLIP \cite{clip}, ALIGN \cite{align}, OpenCLIP \cite{openclip} for the task. Instead of finetuning the VLM on small MLR datasets,  these approaches \cite{scpnet,dualcoop} rely on learning a pair of positive/negative text prompts associated with each class; the prompts associated with a class are learned by maximizing/ minimizing the similarities of their embeddings with those of the embeddings extracted from images containing objects of the class. The additional information captured by the prompts constrains possible matches between the data and the desired labels, in turn limiting the number of parameters needed to be learned. This helps mitigate overfitting on small datasets. At test time, whether a class is present in a given image is ascertained by measuring the similarity of the image embeddings with embeddings extracted from previously learned positive/negative prompts for the class. However, the learning of prompts here is done independently for each class, again neglecting the class co-occurrences.

We propose a two-stage framework that leverages the knowledge of VLMs but injects the co-occurrence information into the models. We obtain an initial estimate of the evidence logits for each class in a subimage in terms of the match between the embeddings of the subimage and those of the positive and negative text prompts associated with the class. Results from the subimages are aggregated to obtain logits for all classes across the image. These independently extracted subimage logits are refined to enforce prior knowledge of the joint probabilities of pairs of classes present in different subimages. Given that a class is present in a training image, the conditional probabilities of other classes being also present are measured from the frequencies of observing those classes in the image. Since the prior probability of a class is known from the VLM output, we can combine it with the conditional probabilities of other classes to obtain joint probabilities of class pairs. Enhancement of the outputs of the independent classifiers by using conditional probabilities is carried out through a graph convolutional network (GCN). GCN thus utilizes the conditional probabilities to improve upon the outputs of the independently obtained class estimates using the vision-language model. As commonly done, we compensate for the differences in the frequencies of different classes occurring in the training images by reweighing the estimated probabilities of different classes to remove the class bias in the training data.

We test our approach on four benchmarks: MS-COCO-small (5\% of the training set), PASCAL VOC, FoodSeg103, and UNIMIB-2016. The first two are commonly used for MLR, whereas we have two additional datasets that have been used in other contexts \cite{unimib2016,foodseg103}. The number of images in each of these datasets is small compared to even many SLR datasets, although MLR calls for larger datasets. This makes MLR here even harder.
Our experiments show that the use of inter-class influence in the second stage of our framework significantly improves performance over the state-of-the-art methods, which detect each class independently. As expected, the advantage is higher for classes for which VLM yields low accuracy but which frequently co-occur with other classes for which VLM accuracy is higher.  
We also show that our loss re-weighing greatly improves the performance on datasets where there is a significant class bias.

\noindent\textbf{Our contributions:}
\begin{itemize}
  \item We propose a two-stage framework to adapt VLMs for MLR with limited annotated data, by enhancing the VLM-based independent class estimates obtained in the first stage, with conditional probability priors extracted from the dataset, using a GCN.
  \item We validate our algorithm quantitatively using mean average precision (mAP) on four MLR datasets. Our method surpasses the previous SOTA MLR approaches by more than 2$\%$ (COCO-14-small), 0.4$\%$ (VOC-2007), 3.9$\%$ (FoodSeg103) and 11$\%$ (UNIMIB2016). 

\end{itemize}

\section{Related Works}
\label{sec:related works}
\subsection{Multi-Label Recognition}

Multi-label recognition is an important, well-studied problem in computer vision, with a wide range of approaches being proposed to tackle it \cite{liu2021emerging}. An important line of work has focused on learning binary classifiers for identifying each class of objects in an image \cite{cole2021multi,misra2016seeing}. These approaches require large labeled datasets for training. They make no use of information that can be derived from modeling the co-occurrence of different classes, which is especially important when only a small amount of annotated data is available for training.

Other works attempt to model label dependencies in an image. \cite{gcnmulti} models label dependencies implicitly by using a deep embedding network shared by classifiers for all different classes of objects. Other works have proposed to use recurrent neural networks (RNNs) to model label dependencies in an image \cite{rnn1,rnn2,rnn4}. Specifically, they cast MLR into a sequence prediction problem, using beam search to find a sequence of objects having the highest likelihood of being present in the image. Like our method, these methods also model label dependencies, but do so via the hidden state of RNNs. However, both these approaches: a deep embedding network or an RNN require large amounts of labeled data for training. 

\cite{gcnmulti} also proposed to guide the training of the deep embedding network using a GCN that explicitly models the label correlation matrix. However, during inference, only the correlation implicitly learned by the deep embedding network is used for MLR. In contrast, we explicitly enforce label correlations using a GCN during both training and inference. We also note that our approach does not rely on implicitly learning label dependencies using the embedding network, as it remains frozen to preserve VLM priors.
\subsection{Vision-Language Models for MLR}
VLMs learn representations that are transferable to a wide range of downstream tasks such as recognition \cite{gest_rec,adapter1,adapter2,adapter3}, retrieval \cite{ret,ret3,ret2}, and segmentation\cite{seg2,open-nerf,csl} by aligning hundreds of millions of image-text(prompts) pairs. Such approaches commonly focus on learning prompts suitable for these downstream tasks \cite{coop}. \cite{dualcoop} adapts VLMs for MLR, proposing to learn only a pair of prompts associated with the presence/ absence of each class while keeping the VLM itself frozen. Text embeddings extracted from the learned prompts are used to gather local evidence from the image features extracted by the VLM, which is then aggregated and combined. SCPNet \cite{scpnet} is another VLM based approach for MLR, differing from \cite{dualcoop} in its use of a GCN to help learn text embeddings for the presence/absence prompts. The GCN models priors derived from class name similarities in CLIP's embedding space, enforcing these priors during prompt learning. \cite{scpnet} also augments training with a self supervised contrastive loss. Both these methods learn independent prompt-based classifiers (as VLM is frozen), not making any use of co-occurrence information during inference. Our method instead learns inter-dependent classifiers by using a GCN to enhance predictions made by the VLM. The inter-dependent classifiers model and make use of conditional probabilities during both training and inference and use co-occurrence information from the actual training dataset.

\subsection{Long tailed Learning}
The distribution of frequency with which objects belong to a class in real-world images often follows a long-tailed distribution\cite{LT4,LT1,LT2,LT3}. Networks trained for multi-class classification on such data tend to perform poorly on the tail classes which have less data available. Several approaches have been proposed to mitigate this issue including data augmentation (augment the tail classes) \cite{LT_aug}, data re-sampling (sample images to obtain balanced distribution) \cite{LT_resamp1,LT_resamp2,LT_resamp3,LT_resamp4}, adjusting classifier margins (classification thresholds vary for every class)\cite{LT_remargin,LT_remargin1} and loss re-weighing  \cite{rewigh2,reweight} being popular. We use loss re-weighing to mitigate the effect of label imbalance on our method when it is trained to model label conditional probabilities. 
%


\section{Method}

Suppose in a given set of images, $ \mathcal{D} = \{ \textbf{x}_{i} \}, i \in \{1 \hdots \lvert \mathcal{D}\rvert \} $, every image $\mathbf{x_{i}}$ may contain objects from up to N classes. The image is thus associated with N  labels $\mathbf{y_{i}} \in \{0,1\}^{N}$ where $y_{i}^{j}$ denotes the presence or absence (1 or 0) of the j-th class in the image. Then the MLR problem requires identification of all labels associated with any input image.

\begin{figure}[t]
    \centering
    \includegraphics[width=1\linewidth]{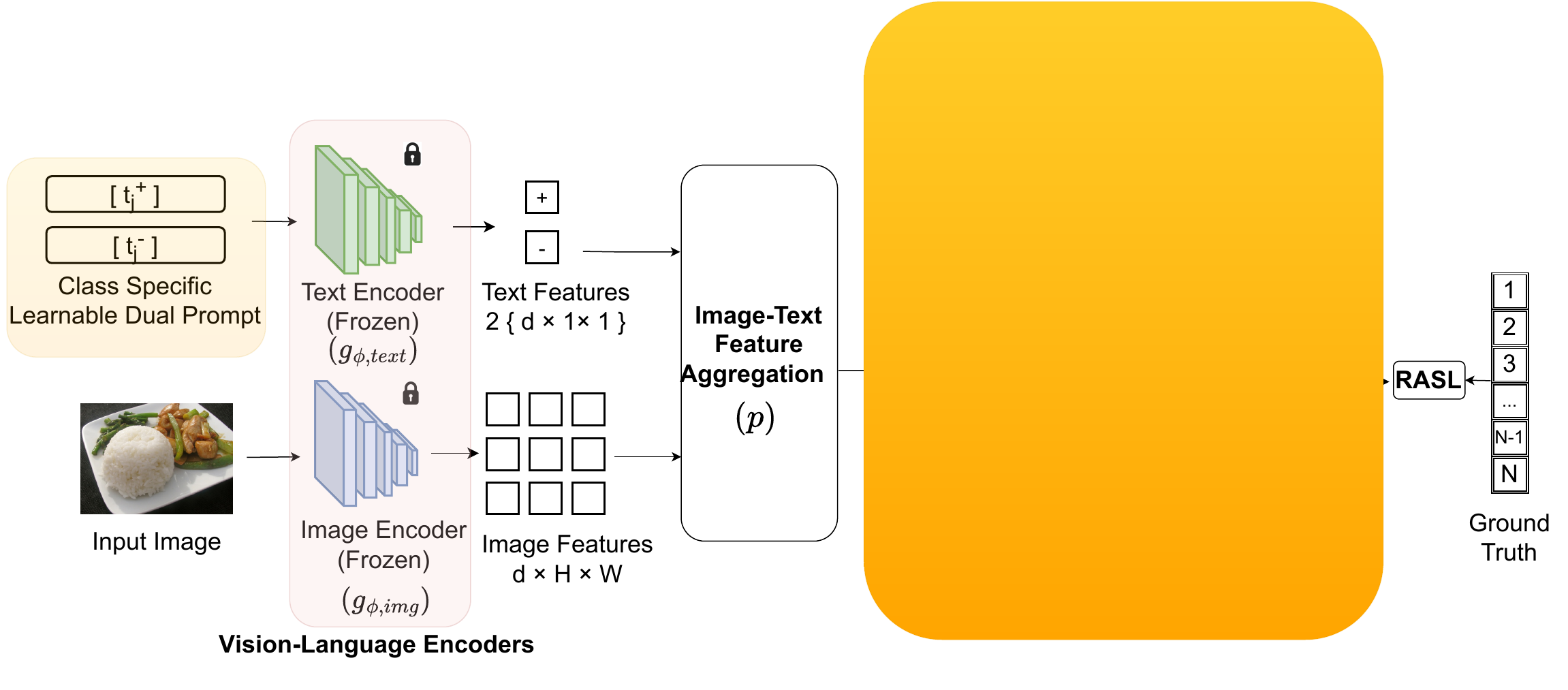}
    \caption{Method Overview: Given an image with multiple objects, we extract image features and text features from the subimages using a vision-language model (CLIP). An image-text feature aggregation module (Sec. \ref{sec: Initial Logits Estimation}) combines these features to identify all classes present in the image as a union of the classes present in the subimages, giving an initial set of image level class logits. These logits are passed to a GCN, that uses conditional probabilities between classes to refine these initial predictions (Sec. \ref{sec:refinement using conditional probability prior}). We train this framework while reweighting the loss generated by classes to address any class imbalance in the training data using a Reweighted Asymmetric Loss (RASL), a weighted version of ASL\cite{asl}.}
    \label{fig:Overview}
\end{figure}

Our approach in this paper uses a VLM  $g_{\phi}$, parameterized using weights $\phi$ (a pair of encoders $g_{\phi,img}$, $g_{\phi, text}$). VLMs are pretrained to align image and textual features over large datasets to learn features suited for various tasks/domains. As mentioned in Sec. \ref{sec:intro}, these models associate a pair of positive and negative text prompts $\{ \mathbf{t_{j,+}}, \mathbf{t_{j,-}} \}$ with each class j (complete set denoted by $\psi$). A text encoder $g_{\phi, text}$ extracts text embeddings from each of these prompts, gives them to an image-text feature aggregation head  $p$, which matches visual features extracted from different subimages with the text embeddings, and combines them to obtain an initial set of logits for each class in the image. We then use a GCN $f_{\theta}$ with weights $\theta$ to refine the logits output by $p$, by leveraging the statistical co-occurrence of classes observed in the training dataset. An overview of our proposed method is given in Figure \ref{fig:Overview}.

The following subsections present the various parts of our method.

\subsection{Initial Logits Estimation}
\label{sec: Initial Logits Estimation}
We use a VLM as a feature extractor and initial classifier for our method. The VLM's image encoder does spatial pooling of windows in the final layer to obtain a single $d$ dimensional feature vector for a single-label image $x_{i}$. This is not suitable for MLR case as spatial pooling operation combines the features of multiple objects in different regions of the image, with overall features being dominated by those extracted from a single object. We remove the pooling layer of the image encoder, using it to get features ${g}_{\phi, img}(\mathbf{x_{i}}) = \mathbf{z_{i}}$ (of shape $d \times H \times W$) for a given image $x_{i}$, hence preserving information from the individual windows.
 

 Our image-text feature aggregation head $p$ is similar to \cite{dualcoop}. For each class $j$, it learns a pair of text prompts $\{ \mathbf{t_{j,+}}, \mathbf{t_{j,-}} \}$, which are projected to d-dimensional embeddings $\mathbf{r_{j,+}}, \mathbf{r_{j,-}}$ using $g_{\phi, text}$. Cosine similarity of the d-dimensional image features at a particular point $(h,w)$ with $\mathbf{r_{j,+}}$ indicates the presence of the class, while similarity with $\mathbf{r_{j,-}}$ indicates its absence. These similarities are aggregated and used by $p$ to give logits $p(\mathbf{z_{i}})$ for the image. 
 

 \subsection{Refining Logits using Conditional Prior}
 \label{sec:refinement using conditional probability prior}
 We refine the logits $p(\mathbf{z_{i}})$ using a GCN, which ensures conformity with the conditional probability priors extracted from the training dataset.
\subsubsection{Estimating Label Conditional Probability Prior:}
To derive the label prior from the dataset, we first calculate the label co-occurrence matrix $C_{N \times N} = (c_{mn})$ over training dataset $\mathcal{D}$. Each entry $c_{mn}$ is given by 
\begin{align}
    c_{mn} = \sum_{i = 0}^{|\mathcal{D}|} y_{i}^{m} \times y_{i}^{n}
\end{align}
The $c_{mn}$ denotes the number of times that objects belonging to classes $m$ and $n$ occur together in an image. Using this, we calculate an estimate of conditional probability matrix $A_{N \times N} = (a_{mn})$, where each entry gives an estimate for the probability $P(y_{i}^{n}= 1 | y_{i}^{m}=1)$:
\begin{align}
    a_{mn} = \hat{P}(y_{i}^{n}= 1 | y_{i}^{m}=1) = \dfrac{c_{mn}}{c_{mm}}
\end{align}



\subsubsection{GCN for Refinement:}

We refine the logits $p_{\psi}(\mathbf{z_{i}})$  using the GCN $f_{\theta}$ which uses the conditional probability matrix $A$ to define the connection weights between its $N$ nodes. Specifically, given that $f_{\theta}$ has L layers, each layer calculates:
\begin{align}
    H^{l} = \rho(AH^{l-1} W^{l})
\end{align} 
where $H^{l-1}$ is the output vector of the previous layer and $\rho$ is a non linearity (Leaky ReLU). $H^{0}$ is defined as the input logits $p_{\psi}(\mathbf{z_{i}})$. 
$W^{l}$ are learnable weights for each layer. Using a GCN layer ensures that the logits are refined using information from only those nodes used for computing the logits, reducing the number of parameters learned while also taking advantage of the conditional probability estimates. After passing through multiple layers of the GCN, we get the updated predictions $f_{\theta}(p_{\psi}(\mathbf{z_{i}}))$. We add the initial logits to the updated logits to obtain our refined logits prediction  $p_{\psi}(\mathbf{z_{i}}) + f_{\theta}(p_{\psi}(\mathbf{z_{i}}))$

\subsection{Training}
\label{sec:RASL}

We train the image-text feature aggregation module $p$ and the GCN $f_{\theta}$, while freezing the VLM $g_{\phi}$. We adopt the widely used Asymmetric Loss (ASL) \cite{asl}, a modified version of the focal loss, to train our network for MLR. 

ASL \cite{asl} addresses the inherent imbalance in MLR caused by the prevalence of negative examples compared to positive ones in training images. Similar to focal loss \cite{focal}, ASL underweighs the loss term due to negative examples. However, it does so using two focusing parameters ($\gamma_{+}$ and $\gamma_{-}$ ) instead of one ($\gamma$) used by focal loss.
However, ASL does not address the issue of sample imbalance, caused by some classes having fewer examples in the dataset. Towards this, we add a loss re-weighting term ($\alpha$) to ASL. Our re-weighed ASL (RASL) is defined as:
\begin{align}
    \mathcal{L}_{RASL}(\hat{y}_{i}^{j}) = \begin{cases}
 & \alpha_{j} \left(1-\hat{y}_{i}^{j}\right)^{\gamma_{+}} \log \left(\hat{y}_{i}^{j} \right) \text{ when } y_{i}^{j} =1 \\
     & \alpha_{j} \left(\hat{y}_{i, \delta}^{j}\right)^{\gamma_{-}}  \log \left(1-\hat{y}_{i, \delta}^{j}\right) \text{ when } y_{i}^{j}=0
\end{cases} \label{eq:instance_RASL}
\end{align}
where $\hat{y}_{i}^{j}$ represents the corresponding prediction associated with label $y_{i}^{j}$; $\hat{y}_{i, \delta}^{j} = \max(\hat{y} - \delta, 0)$, with $\delta$ representing the shifting parameter defined in ASL; 
and $\alpha_{j}$ is the re-weighting parameter for class $j$, defined as:
\begin{align}
    \alpha &= \left( \frac{a_{jj}}{\sum_{j=1}^{N} a_{jj}} \right)^{-1} \label{eq:reweighing}
\end{align}

\noindent Then the total loss over the dataset $|\mathcal{D} |$ is given by:
\begin{align}
    \mathcal{L}_{total} = \sum_{i=1}^{|\mathcal{D}|} \sum_{j=1}^{N} \mathcal{L}_{RASL}(\hat{y}_{i}^{j})
\end{align}



\begin{table}[ht]
\centering
\begin{tabular}{|@{}l|| c | ccccccc@{}|}
\hline
Dataset          & Method          & CP         & CR          & CF1         & OP        & OR        & OF1          & mAP \\ \hline \hline
COCO-small \cite{coco}     & DualCoOp\cite{dualcoop}        & 53.3       & 73.5        & 59.8        & 47.1      & 79.5      & 59.2         & 70.2        \\
                 & SCPNet\cite{scpnet}          & 51.9       & 70.3        & 59.7        & 47.2      & 78.9      & 59.1         & 69.3            \\ \hline
                 & \textbf{Ours }     & \textbf{54.1}       & \textbf{74.3}        & \textbf{62.6}        & \textbf{47.7 }     & \textbf{82.6}      & \textbf{60.5}        & \textbf{72.6 }           \\ \hline \hline

VOC \cite{pascal-voc}             & SSGRL*\cite{SSGRL} & - & - & - & - & - & - & 93.4 \\
 & GCN-ML*\cite{gcnmulti} & - & - & - & - & - & - & 94 \\
 & KGGR*\cite{kggr} & - & - & - & - & - & - & 93.6 \\

& DualCoOp\cite{dualcoop}        & 81.1       & 93.3        & 86.5        & 83.5      & 94.1      & 88.5         & 94.0           \\
                 & SCPNet\cite{scpnet}          & 68.9       & 91.6        & 76.8        & 68.5      & 93.5      & 79.1         & 87.4        \\ \hline
                 &\textbf{Ours} & \textbf{81.1}       & \textbf{94.1}       & \textbf{87.1}       & \textbf{83.6}      & \textbf{94.5}      & \textbf{88.6 }        & \textbf{94.4}        \\ \hline \hline

FoodSeg103 \cite{foodseg103}       & DualCoOp\cite{dualcoop}        & 44.9       & 52.7        & 46.9        & 59.2      & 69.2      & 63.8         & 49.0        \\
                 & SCPNet\cite{scpnet}          & 39.4       & 54.4        & 43.2        & 61.4      & 67.8      & 64.4         & 48.8        \\
                 & Ours w/o reweigh       & 44.8       & 55.0        & 48.0        & 58.6      & 70.2      & 63.9         & 51.3        \\ \hline
                 & \textbf{Ours}  & \textbf{47.1}       & \textbf{55.1 }       & \textbf{50.8}        & \textbf{63.7}      & \textbf{69.9}      & \textbf{66.7}         & \textbf{52.9}        \\ \hline \hline 
UNIMIB \cite{unimib2016}          & DualCoOp\cite{dualcoop}        & 46.9       & 54.7        & 48.4        & 69.0      & 79.0      & 73.7         & 58.1        \\
                 & SCPNet\cite{scpnet}          & 50.5       & 52.9        & 49.9        & 69.6      & 78.4      & 73.8         & 60.0        \\
                 & Ours w/o reweigh       & 52.6       & 59.6        & 53.8        & 73.5      & 83.3      & 78.1         & 64.4        \\ \hline
                 & \textbf{Ours} & \textbf{66.8}       & \textbf{65.8}       & \textbf{64.2}       & \textbf{80.9}      & \textbf{86.1}      & \textbf{83.4}         & \textbf{72.2}        \\ \hline

\end{tabular}
\vspace{6pt}
\caption{Comparison of results obtained by our method and the state-of-the-art baselines, on four MLR datasets in the low data regime: FoodSeg103, UNIMIB 2016, COCO-small (5\% of COCO's training data) and VOC-2007. Our approach achieves the best performance on all metrics: per-class and overall average precisions (CP and OP), recalls (CR and OR), F1 scores (CF1 and OF1), and mean average precision (mAP). * indicates methods that fine-tune the complete backbone network.
}
\label{tab: Low data results}
\end{table}

\section{Experiments}

In this section, we discuss the datasets used, implementation details of our approach, evaluation metrics and provide a thorough analysis of our approach.
 \subsection{Datasets}
We evaluate our method on four different MLR benchmarks: MS-COCO 2014-small and PASCAL VOC 2007, which are widely used MLR benchmarks, as well as FoodSeg103 and UNIMIB 2016, which are smaller MLR datasets suitable for testing in the low data regime. Details of these datasets are given below:
 \noindent\textbf{MS-COCO 2014-small:}  MS-COCO \cite{coco} is another popular MLR dataset and consists of 82,081 training images and 40,504 validation images with objects belonging to 80  classes. To evaluate our methods performance in the low data regime, we use MS-COCO 2014-small, which is a small, randomly selected subset comprising 5\% of MS-COCO 2014 which amounts to 4014 images. During testing, we use the complete validation set.
\\
 \noindent \textbf{PASCAL VOC 2007:} VOC \cite{pascal-voc} is a widely used outdoor scene MLR dataset consisting of 9,963 images from 20 classes. We follow the standard trainval set for training and use the test set for testing.
 \\
\textbf{FoodSeg103: } FoodSeg103 \cite{foodseg103} serves as a benchmark dataset for food segmentation and multi-label food recognition. It consists of 4983 training images and 2135 test images, with a total of 32,097 food instances belonging to  103 different food classes. The number of images per class follows a long-tail distribution typical of real-world datasets. We use the standard train-test data split.
\\
\textbf{UNIMIB 2016:} UNIMIB \cite{unimib2016} is another multi-label food recognition dataset. It consists of 1027 images with 3616 food instances spanning 73 classes. Similar to FoodSeg103, UNIMIB also follows a long-tail distribution typical of real-world datasets. We comply with the official train-test split. 

\subsection{Implementation Details}

In our experiments, we use CLIP (Contrastive Language-Image Pre-Training) \cite{clip} as the VLM. Consistent with recent works that use VLMs for MLR  \cite{Cdul,scpnet,dualcoop}, we select ResNet-101 as the visual encoder and standard transformer within CLIP as the text encoder. Both encoders are kept frozen during our experiments, and we train the GCN and learnable prompts. Following \cite{gcnmulti,scpnet,dualcoop}, we resize the images to $448 \times 448$ for COCO and VOC datasets and to $224 \times 224$ for UNIMIB and FoodSeg103. Similar to previous works \cite{scpnet,dualcoop}  we apply Cutout \cite{cutout} and RandAugment \cite{randaug} to augment training images. We use a 3-layer GCN network for all our experiments. We use SGD for optimizing parameters with an initial learning rate of 0.002, which is reduced by cosine annealing. We train for 50 epochs and use a batch size of 32. We set the loss hyperparameters in Eq. \ref{eq:instance_RASL} as $\gamma_- = 3$, $\gamma_+ = 1$ and $\delta$ = 0.05 . We conduct all experiments on a single RTX A4000 GPU.

\subsection{Evaluation Metrics}
To evaluate the performance of our approach on the four MLR datasets, we use standard metrics, also used by previous MLR approaches \cite{kggr,SSGRL,gcnmulti}. The metrics include the commonly used mean average precision (mAP) as well as class and overall precisions (CP and OP), recalls (CR and OR), and F1 scores (CF1 and OF1). mAP is obtained by calculating the mean of individual average precision (AP) values over all classes. For each class, AP is computed as the area under the Precision-Recall curve. 



\subsection{Results}


We primarily compare our approach  with DualCoOp \cite{dualcoop} and SCPNet \cite{scpnet}, as they are the only other MLR baselines that use VLMs, making them SOTA methods in limited data settings across all four standard benchmarks discussed earlier. As seen in Table \ref{tab: Low data results}, our method outperforms DualCoOp by 0.4\% and SCPNet by  7.0\% mAP on the VOC-2007. On COCO-small, our method outperforms DualCoOp by 2.4\% and SCPNet by 3.3\% mAP. On the FoodSeg103 dataset, our approach significantly improves upon DualCoOp by 3.9\% and SCPNet by 4.1\% mAP. In the UNIMIB dataset, our method achieves substantial performance gains of 14.1\% over DualCoOp and 12.2\% mAP over SCPNet.

Furthermore, for VOC, we extend our comparison to approaches that do not use VLMs and instead rely on complete fine-tuning\cite{kggr,SSGRL,gcnmulti}. These approaches also use a ResNet-101 backbone similar to our visual encoder, but the backbone is initialized with weights pre-trained on ImageNet instead \cite{deng2009imagenet}. Our method also outperforms these methods. More detailed comparison can be found in Table. \ref{tab: Low data results}.


\begin{figure}[ht]
    \centering
        \centering
        \includegraphics[width=0.75\linewidth]{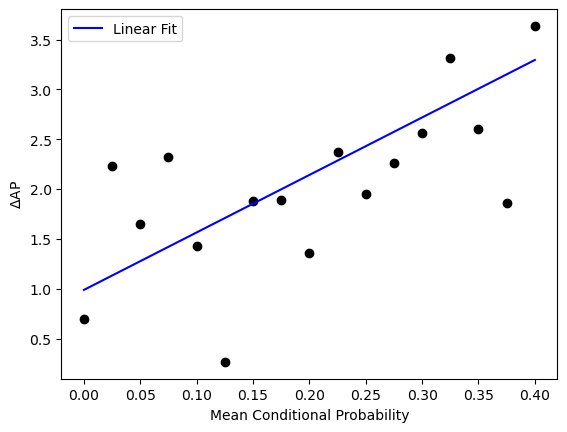}
        \caption{Improvement in average precision ($\Delta$AP) of a class obtained by refining VLM-based initial logits to incorporate the information provided by conditional probabilities, shown as a function of the mean conditional probability of most co-occurring three classes. 
        }
        \label{fig:Corr vs AP}
\end{figure}

\subsection{Impact of the Strength of Conditional Probability on Performance}
\label{Conditional Probability Prior and Performance}
In this section, we determine the impact of the strength of conditional probability of a pair of classes on MLR performance on the COCO-small dataset. Specifically, we observe how the improvement in average precision of a class of objects ($\Delta$AP) brought by our method varies with the average conditional probability of the class paired with the top three other classes it co-occurs the most with. Note that we choose to average the top three values of conditional probabilities of the class because the COCO dataset typically contains an average of around three objects per image. The improvement in ($\Delta $AP) for a class = AP achieved by logits after refinement - AP achieved by the raw VLM logits before refinement.

We visualize the variation in $\Delta$AP with the avg. conditional probability in Figure \ref{fig:Corr vs AP}, where we observe an increasing trend of  $\Delta$AP  with increase in the average of the top-3 conditional probabilities for a given class. Note that for ease of visualization of the scatterplot, we use bins of size 0.02 to group together classes having similar average conditional probabilities. The points represent the average $\Delta $AP value of all classes within the respective bin.

This implies that classes having stronger conditional probabilities with other classes benefit more from our approach of refining logits using conditional probabilities, as is intuitively expected. 

\subsection{Performance on Classes that are Difficult to Recognize}

In this section, we empirically explore the impact of our approach on classes that are difficult to recognize when using image features exclusively. For concreteness, we focus our analysis on the 10 classes in FoodSeg103 and UNIMIB datasets on which the previous state-of-the-art approach (DualCoop\cite{dualcoop}) performs (in terms of CF1) the worst. 

Table \protect\ref{Ablation} compares the performance of our method on these classes with the previous SOTA DualCoOp\protect\cite{dualcoop} and SCPNet\protect\cite{scpnet}. We see that our method significantly improves the performance of these methods, which relies solely on VLMs without modeling any conditional probabilities. Specifically, for DualCoOp, we observe a growth of 22.3\% in CP, 33.9\% in CR and 34.8\% for CF1 on UNIMIB2016, and 15.6\% in CP, 7.2\% in CR and 11.89\% in CF  on FoodSeg103. For SCPNet, we see gains of 27.1\% CP, 25.2\% CR, and 26.5\% CF1 on UNIMIB2016, and 15.8\% CP, 5.8\% CR, and 12.4\% CF1 on FoodSeg103.

This underscores the importance of the information obtained by modeling joint class probabilities in recognizing classes of objects that are difficult to recognize from image features alone.

\begin{table*}[htbp]
    \centering
    \setlength{\tabcolsep}{8pt} 
    \small 
    \begin{tabular}{|l|c|c|c|c|c|c|c|c|}
        \hline
        & \multicolumn{3}{c|}{UNIMIB} & \multicolumn{3}{c|}{FoodSeg103} \\
        \hline
        Methods & CP & CR & CF1  & CP & CR & CF1  \\
        \hline
        DualCoOp & 25.4 & 26.2 & 24.3  & 13.7 & 19.7 & 16.5  \\
        SCPNet & 30.5 & 34.8 & 32.5  & 12.9 & 21.1 & 16.0  \\
        Ours w/o reweigh & 41.9 & 57.5 & 44.9 & 14.8 & 22.5 & 18.7  \\
        Ours & \textbf{57.6} & \textbf{60.0} & \textbf{59.1} & \textbf{28.7} & \textbf{26.9} & \textbf{28.4} \\
        \hline
    \end{tabular}
    \vspace{6pt}
    \caption{A comparison of the average performance of our approach with the previous state-of-the-art VLM-based method DualCoOp\cite{dualcoop} and SCPNet\cite{scpnet} on classes that are difficult to recognize using only visual features (having 10 lowest CF1 values on the FoodSeg103\cite{foodseg103} and UNIMIB\cite{unimib2016}). Our approach significantly improves MLR performance on such classes due to its use of information derived from class conditional probabilities.}
    \label{Ablation}
\end{table*}

 \subsection{Effect of Loss Reweighing}
 
In this subsection, we investigate the effects of our loss reweighing strategy for UNIMIB2016 and FoodSeg103, which exhibit significant class imbalance. We analyze its impact on performance on  (1) All classes as a whole and (2) Classes that are difficult to recognize using only visual features, and hence have a greater reliance on information obtained from conditional probabilities.


(1) \textbf{All classes as a whole:} As observed in Table \ref{tab: Low data results}, loss re-weighing improves the performance of our method by 1.6\% and 7.8\% in mAP on FoodSeg103 and UNIMIB2016, respectively. 

This demonstrates the utility of using loss re-weighing when training a method modeling the joint probability distribution of classes as opposed.


(2) \textbf{Classes that are difficult to recognize using only visual features:} As seen in Table \ref{Ablation}, loss re-weighing also significantly boosts the performance of our method on these classes. It improves CP, CR and CF1 by 13.9\%, 4.4\% and 9.7\%, respectively on the FoodSeg103 dataset and by 15.7\%, 4.5\% and 14.2\%, respectively on the UNIMIB2016 dataset. Note that these increases are higher than corresponding gains observed for all classes in the dataset, signifying that loss re-weighing delivers larger benefits to classes more reliant on conditional probabilities for recognition (as opposed to those that derive more information from initial logits estimated from logits)

\section{Conclusion}
In this paper, we present a novel two-stage framework for multi-label recognition when only a small number of annotated images are available. Our framework builds on recent methods that make use of VLMs to counter this paucity of labeled data but overlook information derived from co-occurence of object pairs. Our framework refines the logit predictions made by VLMs adapted for multi-label recognition by leveraging known conditional probabilities of class pairs derived from the training data distribution. Specifically, we use a graph convolutional network to enrich the logits predicted by the VLM with information from conditional probabilities of classes. Our method outperforms all state-of-the-art approaches on 4 MLR benchmarks: COCO-14-small, VOC 2007, FoodSeg103 and UNIMIB2016 in a low data regime, demonstrating the utility of modeling class co-occurrence in such cases.


\section{Limitations}
(1) If the independent classifiers learned by state-of-the-art approaches (relying on only visual information, not modeling the conditional probability of class pairs) are strong, and characterized by a high average precision (AP) of each class, our method would yield lower improvements. However, in practice, many MLR datasets are not very large, with independent classifiers learned from them being relatively weak. MLR on such datasets is likely to benefit significantly from our method. \\
(2) As shown in Figure. \ref{fig:Corr vs AP}, the advantage provided by our method is higher when the conditional probability of pairs of classes co-occurring in an image is higher. For images that consist of objects which are rarely found together, our method provides very little added benefit over independent classifiers.  

\section{Acknowledgement}\vspace{-5pt}
We thank Kamila Abdiyeva for her insightful feedback on the manuscript. The support of the Office of Naval Research under grant N00014-20-1-2444, of USDA National Institute of Food and Agriculture under grant 2020-67021-32799/1024178 and Vizzhy.com are gratefully acknowledged.

\bibliographystyle{splncs04}
\bibliography{main}
%




\end{document}